\documentclass[10pt,twocolumn,letterpaper]{article}

\usepackage{cvpr}              

\definecolor{cvprblue}{rgb}{0.21,0.49,0.74}
\usepackage[pagebackref,breaklinks,colorlinks,allcolors=cvprblue]{hyperref}

\def\confName{CVPR}
\def\confYear{2025}

\def\paperID{****} 
\title{Towards Chunk-Wise Generation for Long Videos}

\author{Siyang Zhang\\
University of Central Florida\\
{\tt\small si122915@ucf.edu}
\and
Ser-Nam Lim\\
University of Central Florida\\
{\tt\small sernam@ucf.edu}
}

\begin{document}
\twocolumn[{%
\renewcommand\twocolumn[1][]{#1}%
\maketitle
\begin{center}
    \vspace{-20pt}
    \includegraphics[
    width=0.8\textwidth,
    height=0.31\textwidth
    ]{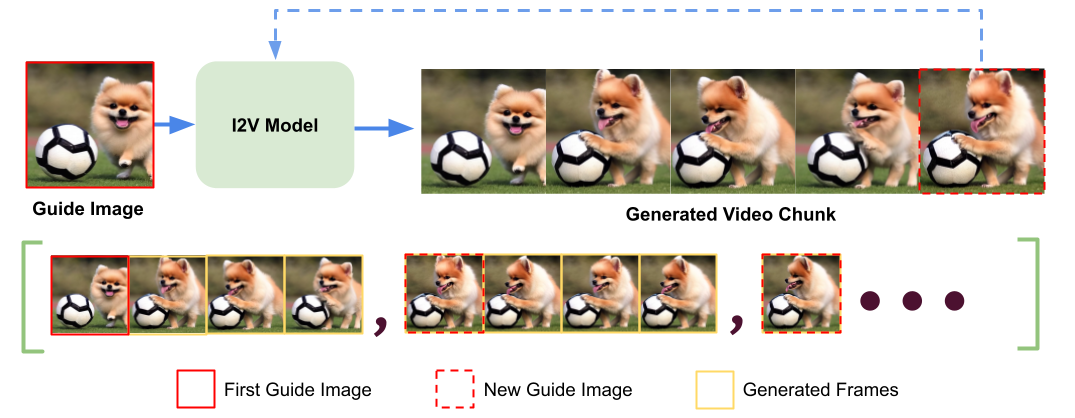}
\vspace{-0.4cm}
\captionof{figure}{\footnotesize
An overview of the pipeline of autoregressive chunk-by-chunk long video generation. Each time an Image-to-Video(I2V) model takes in a guide image as condition and generate a short video chunk. Then, the model will take the last frame of collected videos, and predict the next video chunk.
}
    \vspace{3pt}
    \label{fig:motivation}
    \end{center}%
}]



\begin{abstract}
Generating long-duration videos has always been a significant challenge due to the inherent complexity of spatio-temporal domain and the substantial GPU memory demands required to calculate huge size tensors. While diffusion based generative models~\cite{ho2020denoising} achieve state-of-the-art performance in video generation task, they are typically trained with predefined video resolutions and lengths. During inference, a noise tensor with specific resolution and length should be specified at first, and the model will perform denoising on the entire video tensor simultaneously, all the frames together. Such approach will easily raise an out-of-memory (OOM) problem when the specified resolution and/or length exceed a certain limit. One of the solutions to this problem is to generate many short video chunks autoregressively with strong inter-chunk spatio-temporal relation and then concatenate them together to form a long video. 
In this approach, a long video generation task is divided into multiple short video generation subtasks, and the cost of each subtask is reduced to a feasible level. 
In this paper, we conduct a detailed survey on long video generation with the autoregressive chunk-by-chunk strategy. We address common problems caused by applying short image-to-video models to long video tasks and design an efficient $k$-step search solution to mitigate these problems.

\end{abstract}    
\section{Introduction}
\label{sec:intro}

Diffusion models have achieved state-of-the-art in vision tasks since they were firstly introduced for image generation in 2020~\cite{ho2020denoising}. Unlike previous model structure such as GAN~\cite{goodfellow2014generative}, they perform an iterative denoising process on a Gaussian noise latent, removing noise step-by-step and finally resulting in a clear tensor. Such an iterative denoising inference strategy has demonstrated robustness and diversity in the generated videos. However, unlike autoregressive models~\cite{yu2023magvit, yu2023magvit2, kondratyuk2023videopoet, wu2022nuwa, sun2024autoregressive} that predict tokens one by one based on previous tokens, diffusion models must work on the \emph{entire} noisy latent tensor with a predefined size. 
This cost is rather feasible and acceptable in image domain since an image has fixed resolution, and most image resolutions varies within the range of $32\times 32$ to 2K. However, videos can vary from short GIFs with only a few seconds, to long films or documentaries with couples of hours, 
creating significant memory requirement both during training and inference. Although in theory the video latent tensor size can be customized to a desired longer length after training with a smaller tensor size to save memory, we are faced with a
train-inference discrepancy in video length. That is, if a video diffusion model is only trained with short videos, it can hardly generate a much longer video without a significant degradation in quality during inference by only naively setting a larger initial noise tensor shape, assuming that there are sufficient memory to do so. 

A video is a sequence of images that share strong spatio-temporal relations. Such inherent property allows the possibility to perform autoregressive chunk-by-chunk video generation, by conditioning the generation of a video chunk on previously generated video chunks. An intuitive implementation is to leverage an Image-to-Video (I2V) diffusion model that iteratively accepts a guide image as conditioning input, and outputs a short chunk that follows the guide image. Doing so alleviates the memory requirement that would otherwise be insurmountable if we attempt to generate a long video all at once. An alternate approach could be autoregressive methods~\cite{yu2023magvit2, kondratyuk2023videopoet, wu2022nuwa}, that generate videos token by token, and are thus not faced with the same memory constraint that diffusion models face. Unfortunately, most of these methods are either not open sourced or have yet to demonstrate the same level of efficacy as video diffusion.

\begin{figure}
    \centering
    \includegraphics[width=\linewidth]{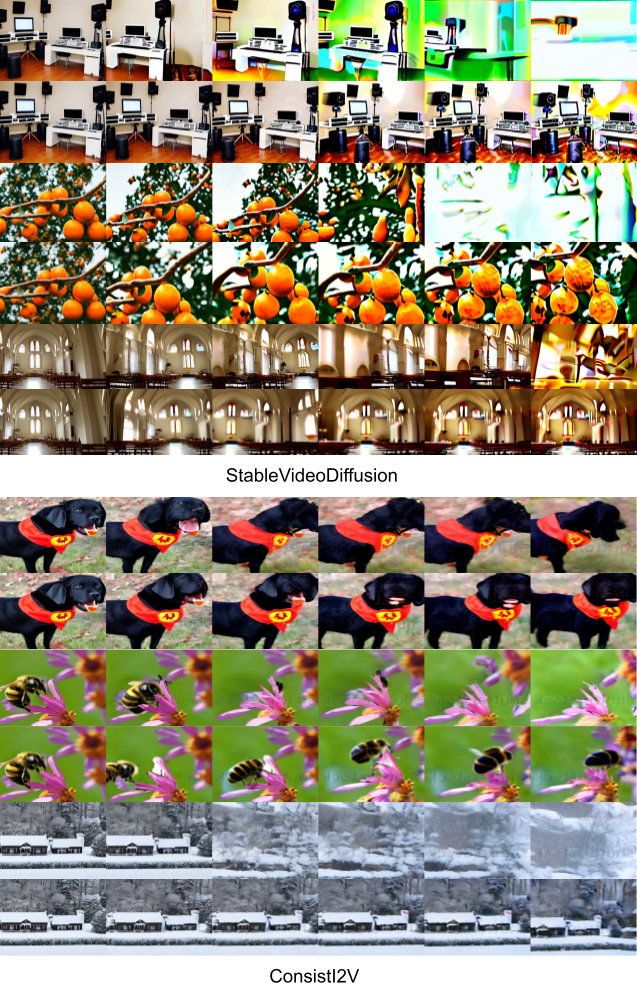}
    \caption{\footnotesize
    Examples of degradation effect on video quality in chunk-by-chunk video generation by StableVideoDiffusion~\cite{blattmann2023stablevideodiffusion} and ConsistI2V~\cite{ren2024consisti2v}.  
    For each guide image, we perform naive chunk-by-chunk generation (top row) and $k$-step search generation (bottom row). The model created some artifacts in each chunk, and the cumulated effect will at last destroy the long video as the number of chunks increases. Our $k$-step search helps to mitigate the degradation.}
    \label{fig:svd_examples}
\end{figure}

\begin{figure}
    \centering
    \includegraphics[width=1.0\linewidth]{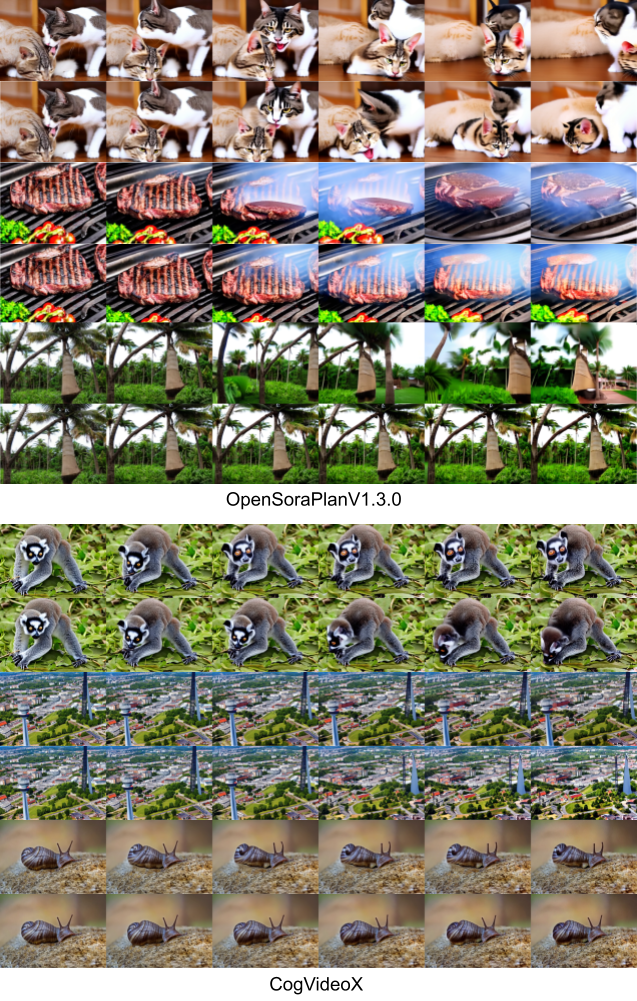}
    \caption{\footnotesize
    Examples of long videos generated by OpenSoraPlanV1.3.0 and CogVideoX. For each guide image, we perform naive chunk-by-chunk generation (top row) and $k$-step search generation (bottom row). These models are more robust to initial noise.}
    \label{fig:cog_osp}
\end{figure}

However, applying a pretrained I2V diffusion model to a long video generation task in a chunk-by-chunk manner requires a strong assumption that the short video generated at each iteration remains of high quality so as not to misguide the generation of the next chunk. Unfortunately, in our observation, that assumption is rarely true because models are not perfect, and that fact leads to some challenge such as frame quality degradation. We especially observed that smaller models with simpler motion modules are extremely vulnerable to degradation as the number of chunks increases (see Figure~\ref{fig:svd_examples}). Here, we observed that the initial noise plays a significant role in the generation. Some initial noise will lead to a local high-quality chunk with more consistent object and smoother motion while some will lead to a local bad-quality chunk that contains degradations such as color shifting and object distortion. If we happen to sample a bad noise along the line, errors will accumulate down the line and the degradation worsens. On the other hand, large I2V models such as OpenSoraPlanV1.3.0~\cite{yuan2024opensoraplan} and CogVideoX~\cite{yang2024cogvideox} are usually more robust to the initial noise (See Figure~\ref{fig:cog_osp}), as a trade-off to larger model size, more parameters, much more expensive training and inference cost. 

Based on this observation, an intuitive solution is to discern the quality of the initial noise tensor. However, modeling a Gaussian noise is hard, and a ``good" noise is not necessarily good for every conditioning inputs. Some existing noise initialization method~\cite{wu2025freeinit} tried to refine the low-frequency component of an initial Gaussian noise by first fully denoising it through the base model, conducting the forward diffusion to re-add a new Gaussian noise to a fully noisy level, and then composing a refined noise by adding up the low-frequency component of the re-diffused latent and high-frequency component of a new random noise through Fast-Fourier-Transform. However, such algorithm demands many runtimes of full denoising, 
which is extremely expensive and slow. 

In this paper, we propose a fast evaluation method that only takes $k$ denoising steps to rapidly generate a video that is suboptimal but sufficient for evaluating the quality of the noise. Even though the suboptimal video is of low visual quality, it still captures the overall layout of its fully-denoised counterpart, and thus helpful in deciding whether an initial noise is good. Our main contributions are: 
\begin{enumerate}
    \item This paper offers a detailed analysis of utilizing several widely used I2V models for training-free chunk-by-chunk long video generation, including an analysis of the impact of initial noise on per-chunk video quality. Chunk-by-chunk generation has the potential to overcome many of the efficiency problems associated with long video generation; while previous work~\cite{ren2024consisti2v, chen2024controlavideocontrollabletexttovideodiffusion} has briefly touched on this topic, there has been no in-depth analysis conducted on it.
    \item We propose an efficient search method that evaluates and selects the best noise during generation, helping to mitigate the effect of bad initial noise. This is especially instrumental for small I2V models, which is more prone to error accumulations in chunk-by-chunk generation, but yet are much more efficient for practical purposes. On the other hand, we also show that larger I2V models such as OpenSoraPlanV1.3.0~\cite{yuan2024opensoraplan} and CogVideoX~\cite{yang2024cogvideox} do not need much intervention and a naive chunk-by-chunk suffices, but at the cost of a much higher inference lapse.
    \item We demonstrate that autoregressive chunk-by-chunk video generation (whether with our bad noise mitigation or simply naive) is a promising method for long video generation task by supporting the idea with strong and insightful empirical results. More importantly, this work opens the door for a practical and easy to use paradigm to generate long videos, which can be easily applied to future video generation techniques.
\end{enumerate}

\section{Related Works}
\label{sec:formatting}

Generative models for vision tasks have evolved from GANs~\cite{goodfellow2014generative} to diffusion models~\cite{ho2020denoising}. One of the most widely known application is Text-to-Image (T2I), leveraging a UNet~\cite{ronneberger2015unet} structure to model the reverse process of adding Gaussian noise to clear latents. Based on these well developed T2I models' capability of understanding and handling 2D image knowledge, further application such as Text-to-Video unleashes the possibility for creative works in a more complex video domain.

\subsection{Text-to-Video (T2V) Diffusion Models}
Similar to Text-to-Image (T2I) models, T2V models accept textual embedding of raw text prompt by some pretrained text encoder such as CLIP~\cite{radford2021clip} or T5~\cite{raffel2020t5}, via cross attention mechanism, enabling meaningful video generation that corresponds to the input text prompt. Early works such as AnimateDiff~\cite{guo2023animatediff} typically leverage a pretrained 2D UNet~\cite{ronneberger2015unet, ho2020denoising} for T2I task, and introduce additional modules~\cite{guo2023animatediff, wang2023modelscope, chen2024videocrafter2} that monitor motion features across frames, to each UNet block. These extra modules are mostly implemented with 1D convolution and self-attention mechanism, together with 2D UNet, to compose a 2D+1D, a.k.a. pseudo 3D UNet. Later works such as OpenSoraPlan~\cite{yuan2024opensoraplan} and CogVideoX~\cite{yang2024cogvideox} adopt Diffusion Transformer (DiT)~\cite{peebles2023dit} that supports 3D attention, as backbone structure, and demonstrate better performance in visual quality including less flickering and more smoothness.

\subsection{Image-to-Video (I2V) Diffusion Models}
Unlike T2V, I2V models accept image conditioning and animate that conditioning image to result in a video. Also, text prompt is usually abstract but image guidance is explicit, and thus it is expected that the output video of an I2V model should be visually consistent to the guide image. Due to such a requirement, in actual implementation, the guide image serves as a strong condition, and is usually expanded on the time dimension and then concatenated to the video tensors in channel dimension (StableVideoDiffusion~\cite{blattmann2023stablevideodiffusion}), together with a masked condition (OpenSoraPlan-I2V~\cite{yuan2024opensoraplan}) needed by the self-attention mechanism. This is enabled by introducing extra channels to the first convolution layer in the diffusion model. 

\subsection{Long Video Generation}
Diffusion based video generation models perform iterative denoising steps on an randomly initialized Gaussian noise that is of the same size as the expected video output. Such design brings challenges for long video generation. One of the more severe problems is the infeasibility of generating a very long video by initializing a noise tensor as big as the target video size, as this will result in GPU memory requirements that exceed the available hardware. Instead, approaches that generate a small chunk at a time, followed by connecting the chunks while ensuring visual connectivity have been proposed. There are two types of such approach.
 
 \subsubsection*{First-In-First-Out (FIFO) Generation}
FIFO-diffusion~\cite{kim2024fifo} leverages a pretrained T2V diffusion model and reschedules the denoising strategy from a simultaneous denoising on all frames with the same noise timestep to a progressive denoising on each frame with increasing noise timesteps. Although this method is as memory efficient as the base T2V inference, it has limitations such as train-inference discrepancy between fixed and progressive timestep schedules. Furthermore, due to the intrinsic abstract characteristic of text prompt and stochasticity of T2V model, object consistency may suffer as the video length increases. 

 \subsubsection*{Autoregressive Generation}
 Autoregressive generation utilizes previously generated results to predict next result in a sequential manner. There are works such as MagVit-V2~\cite{yu2023magvit2} that tokenize videos and allow next video token prediction, enabling long video generation. However, these works are not open sourced and thus their performances in long video generation remain unclear. 
 
 Another approach that we name chunk-by-chunk generation is designed to denoise only a single video chunk in each runtime, which is then repeated using the latest video frame collected so far as the guide image. This approach may face quality degradation due to the accumulation of unexpected artifacts generated in each chunk. We observed that a major cause of these artifacts are the noise input to the unet. A ``bad" noise tensor would result in significant errors that are hard for the diffusion model to recover from. There are works such as FreeInit~\cite{wu2025freeinit} and FrameInit~\cite{ren2024consisti2v} addressing this problem through an expensive and time-consuming procedure that includes refining initial noises by Fast-Fourier-Transformation, low-high frequency components decomposition, together with full step denoising and re-diffusing.

\begin{figure*}
\begin{center}
    \vspace{-20pt}
    \includegraphics[width=1.0\textwidth]{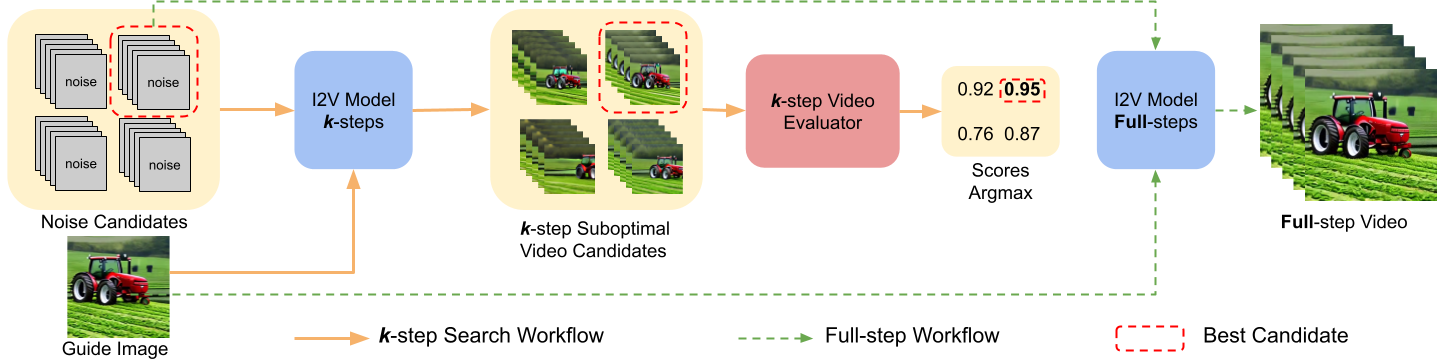}

\caption{\footnotesize
$k$-step search: we first prepare $m$ initial noises and then for each of them, call the base I2V model to only denoise for $k$ steps, resulting in $k$ suboptimal short video candidates. After that, we explicitly evaluate the $k$ video candidates and find the one with the best quality. Finally, we use the noise that leads to the best video to perform a full step denoising.
}
    \label{fig:kstep_search}
    \end{center}%
    \end{figure*}
\section{Method}

In this section, we will first briefly introduce the preliminaries of diffusion based video generation models, analyzing the main obstacle for these models in the task of long video generation. Then, we will elaborate our simple, efficient and training-free approach that utilizes pretrained I2V models on long-video generation task.

\subsection{Preliminaries}
Diffusion models learn a backward process to reverse a procedure that iteratively adds Gaussian noise to clear inputs, a.k.a., the forward process. In a pre-designed variance schedule $\beta s = [\beta_0, ...,\beta_{T-1}]$, such that $0 < \beta_i < \beta_j < 1$ for any $i<j$, and $T$ is the number of predefined number of diffusion steps. a forward diffusion will iteratively add random Gaussian noises to a clear input, to get a noisy version of the input. Then, the model is trained to be able to predict the noise added at current noise level, and remove that noise to recover a less noisy latent.
In a forward diffusion process, there are $T$ steps (usually $T$=1000)~\cite{ho2020denoising}. If we have a clear latent $z_0$, then a noisy version $z_t$ of $z_0$ satisfies the following distribution:
\begin{equation}
    q(z_t|z_0) = \mathcal{N}(z_t;\sqrt{\Bar{\alpha}_t}z_0, (1-\Bar{\alpha}_t)\mathbf{I}),
\end{equation}
where $\alpha_t = 1 - \beta_t$, $\Bar{\alpha_t} = \prod_s^t \alpha_s$, and $\textbf{I}$ is an all-one tensor with same shape as $z_0$.
Thus, in actual implementation, we have:
\begin{equation}
    z_t = \sqrt{\Bar{\alpha}_t}z_0 + \sqrt{1-\Bar{\alpha}_t}\epsilon, \epsilon \sim \mathcal{N}(0, \mathbf{I}).
\end{equation}
In an iterative denoising inference stage, the video diffusion model takes in a pure Gaussian noise that is of the same shape as the input training video latents. Hence, during one runtime of a denoising loop, only the content of the latents will become more clear, but the resolution and duration are always fixed. Even though some models are trained on videos with various resolution and duration, and allow various input tensor shape in inference time, requiring a larger resolution and/or a longer duration means more memories are required and thus may trigger an out of memory (OOM). 

\subsection{Autoregressive Generation with I2V}
To avoid the OOM problem for long video generation, it is intuitive to adapt an image-to-video (I2V) model autoregressively. 
%


    
%
By naively running an I2V model for $n$ runtimes, each time setting the guide image to be the last frame of the previous video chunk, we are able to collect a video as long as possible, without running into memory problem. However, this algorithm is flawed. For every video chunk generation, a random noise is sampled, and the output strongly depends on the initial noise, and the last frame of the previous video chunk. When a bad noise is sampled, a bad chunk will be generated, leading to a bad guide frame for the next chunk, resulting in deterioriating quality as the number of chunks increases.

\subsection{Initial Noise Evaluation}

Like image diffusion videos, the learning objective for video diffusion models is:
\begin{equation}
    \min_\theta \mathbb{E}_{z_0, t, c} \left\| \epsilon - \epsilon_\theta(z_t, t, c) \right\|_2^2,
\end{equation}
where $z_0$ is a clear latent, $t$ is a timestep between [0, 999], $c$ is the conditioning input, $z_t$ is a noisy version of $z_0$, $\epsilon \sim \mathcal{N}(0, \textbf{I})$, and $\epsilon_\theta$ is a denoising model with learnable parameters $\theta$.
In the inference stage, a pure Gaussian noise $z_T = \mathcal{N}(0, \mathbf{I})$ is randomly sampled and regarded as a level-1000 noisy inputs. Then the diffusion model will iteratively predict the current noise to be removed, and update $z_T$ until reaching the last step.

However, there exists a train-inference discrepancy~\cite{wu2025freeinit} for the input noisy latent. In training, the input noisy latent is obtained from a clear ground truth latent, with some noise level $t$. On the other hand, during inference, an input noise is directly sampled from a pure Gaussian distribution. Even though with the largest possible $T=999$, the resulting training noisy latent does resemble a Gaussian noise, the actual distribution of $T=999$ noisy latent is inherently not exactly the same as a pure Gaussian distribution. To be specific, in training, the space of most noisy latents is $S_{\text{train}} = \sqrt{\Bar{\alpha}_{T}} z + \sqrt{1 - \Bar{\alpha}_{T}} \epsilon$,
where $\epsilon \sim \mathcal{N}(0, \mathbf{I})$, $\Bar{\alpha}_t = \prod_{i=1}^t \alpha_i = \prod_{i=1}^t (1 - \beta_i)$, $\beta$ is the predefined variance schedule and $z$ is in the clear image latent space. On the other hand, during inference, the latent space is $S_{\text{inference}} \sim \mathcal{N}(0, \mathbf{I})$.
Even though $\sqrt{\Bar{\alpha}_{T}}$ is very close to 1, $S_{\text{train}}$ and $S_{\text{inference}}$ are not equivalent. It is therefore no wonder that we observed in our experiments that even with the same input condition, different initial Gaussian noises will lead to different output videos, with a huge variations in quality (See Figure~\ref{fig:differet_init_noise}). This is a result of the discrepancy between the initial noises sampled during training and inference, and those initial noises during inference that happen to be similar to the ones seen by the model during training tend to generate higher quality outputs. 
Hence, we argue that, if we are able to find a good initial noise every time we sample a video chunk, the cumulative worsening effect will be mitigated.

\subsection{Brute Force and $k$-step Search}
In previous observation, we noticed that the choice of initial noises significantly determines the visual quality of generated videos. However, since the initial noises are pure Gaussian, it is hard to find an analytic evaluation metric to the noises. Further, we observed that a noise that produces a high quality video given a guide image and a model, is not necessarily good in the context of another guide image or model. Thus, evaluation based solely on the noise itself will not in principle lead to the correct choice. 



Fortunately, the denoising model is innately capable of telling us whether a noise is good based on the generated video. Here, a brute force search can be formulated as a quality check by comparing the quality of different generations based on different noises and selecting the best candidate. We observed that this greatly improves the quality of long videos of multiple chunks, especially for mitigating the cumulative worsening effect. However, it requires many full step inferences to find a good video chunk, which is extremely expensive. For example, if sampling a video from pure noise takes 100 DDIM~\cite{song2020ddim} steps, then by applying brute force on 10 candidate noises, we will need a total of 1,000 steps, which is prohibitively too expensive. 


To handle this problem, we propose a $k$-step evaluation strategy (See Figure~\ref{fig:kstep_search}), which only takes $k$ steps of sampling rather than a full inference to generate videos that, even though are suboptimal, are sufficient for the evaluation and selection of noise.
Most diffusion models that are trained based on a DDPM~\cite{ho2020denoising} scheduler have predefined 1000 steps, but utilize advanced schedulers during inference such as DDIM~\cite{song2020ddim}, PNDM~\cite{liu2022pndm} and EulerAncestralDiscrete~\cite{karras2022eulerdiscreate}. These schedulers allow for customized number of sampling steps, trading off between the output quality and the number of sampling steps, with the most commonly adopted number of steps ranging from 25 to 100. Further, since the stochasticity of diffusion model mostly comes from the initial noise (except for some schedulers~\cite{song2020ddim, karras2022eulerdiscreate, lu2022dpmsolver} that introduce minor randomness during each sampling step, and this can also be controlled by fixing a random seed), we are able to repeat the output of different runtime by setting the same initial noise. This means that given the same initial noise, the proposed $k$-step evaluation is able to indicate the overall quality of an otherwise fully denoised output.

In practical implementation, $k$ can be very small compared to the full sampling step. For example, when the number of steps for full sampling is 100, $k=5$ and $m=10$, the total number of extra $k$-step is 50, as opposed to 900 (1000-100) in the case of brute force.


\subsection{$k$-step Evaluation}
Videos generated with $k$-step are obviously suboptimal compared to fully denoised videos. As such, the choice of the evaluation metric used to select the best noise needs to be carefully chosen. In principle, the metric should not be sensitive to pixel-wise low-level details, and regular metrics that capture low-level details might not be effective indicators. We found empirically that the cosine similarity between the CLIP embeddings~\cite{radford2021clip} of each frame to the guide image to be an effective indicator, because CLIP will encode an image to a latent that captures high-level features. Specifically, for a video $V$ of length $T$ generated with $k$-step, the evaluation score assigned to it with respect to the guide image $I_{guide}$ becomes, $V[j]$ being frame $j^{th}$ of $V$:
\begin{equation}
    \text{min}\{ 
    \text{CosSim}(\text{CLIP}(I_{guide}, V[j])); 0 \leq j < T \}
\end{equation}

\begin{algorithm}
\caption{Autoregressive $k$-step Search}\label{alg:kstep}
\begin{algorithmic}

\State $D \gets \text{Denoising model}$
\State $f \gets \text{An evaluator on suboptimal Videos}$
\State $g \gets \text{Guide image}$
\State $c \gets \text{Other conditional inputs}$
\State $k \gets \text{Number of sampling steps for suboptimal evaluation}$
\State $s \gets \text{Number of full sampling steps}$
\\
\State $vs = [\ ]$
\For{i in [1, 2, ..., n]}
    \State $\epsilon s = [\ ]$
    \State $v_{cand} = [\ ]$
    \For{j in [1, 2, ... m]}
        \State $\epsilon_j \sim \mathcal{N}(0, \mathbf{I})$
        \State $\epsilon s.add(\epsilon_j)$
        \State $v_j = D(\epsilon_j, c, g, num\_step=k)$
        \State $v_{cand}.add(v_j)$
    \EndFor
    \State $best\_idx = \text{Argmax}(f(v_{cand}))$
    \State $\epsilon^* = \epsilon s[best\_idx]$
    \State $v = D(\epsilon^*, c, g, num\_step=s)$
    \State $g = v[-1]$
    \State $vs.add(v)$    
\EndFor
\State \Return $concat(vs)$
\end{algorithmic}
\end{algorithm}

\begin{table*}
\small
\centering
\footnotesize

\scalebox{0.78}{

\begin{tabular}{lcccccccccccccccc}
\toprule
{\textbf{Methods/Metrics}} &  
\multicolumn{4}{c}{\textbf{Subject Consistency$\uparrow$}} &  \multicolumn{4}{c}{\textbf{Background Consistency$\uparrow$}}  &  \multicolumn{4}{c}{\textbf{Temporal Flickering$\uparrow$}} & \multicolumn{4}{c}{\textbf{Motion Smoothness$\uparrow$}} \\
\midrule
Number of Chunks   

&  min$\uparrow$ & max$\uparrow$ & range$\downarrow$ & std$\downarrow$ 
&  min$\uparrow$ & max$\uparrow$ & range$\downarrow$ & std$\downarrow$
&  min$\uparrow$ & max$\uparrow$ & range$\downarrow$ & std$\downarrow$
&  min$\uparrow$ & max$\uparrow$ & range$\downarrow$ & std$\downarrow$

\\
\cmidrule(lr){1-1} \cmidrule(lr){2-5} \cmidrule(lr){6-9} \cmidrule(lr){10-13} \cmidrule(lr){14-17}

StableVideoDiffusion 
& 0.8485 & 0.9638 & 0.1153 & 0.0344 
& 0.9091 & 0.9707 & 0.0616 & 0.0182 
& 0.8585 & 0.9208 & 0.0622 & 0.0187 
& 0.9014 & 0.9662 & 0.0648 & 0.0198
\\

ConsistI2V 
& 0.9432 & 0.9876 & 0.0443 & 0.0136 
& 0.9597 & 0.9899 & 0.0301 & 0.0094 
& 0.9426 & 0.9777 & 0.0351 & 0.0111 
& 0.9635 & 0.9824 & 0.0189 & 0.0058
\\

OpenSoraPlanV1.3.0 
& 0.9384 & 0.9548 & 0.0164 & 0.0050 
& 0.9593 & 0.9710 & 0.0118 & 0.0037 
& 0.9649 & 0.9713 & 0.0064 & 0.0021 
& 0.9862 & 0.9879 & 0.0018 & 0.0006
\\

CogVideoX 
& 0.9663 & 0.9794 & 0.0131 & 0.0041 
& 0.9758 & 0.9858 & 0.0100 & 0.0031 
& 0.9754 & 0.9835 & 0.0081 & 0.0025 
& 0.9881 & 0.9915 & 0.0034 & 0.0011
\\

\bottomrule

\end{tabular}

}

\caption{\footnotesize
We randomly sample 10 different initial noises for every guide image, generates 10 single-chunk videos and evaluate them with VBench metrics. All values are averaged among all test samples.
}
\label{tab:noise_analysis}
\end{table*}

\section{Experiments}

\begin{figure}
    \centering
    \includegraphics[width=0.5\textwidth]{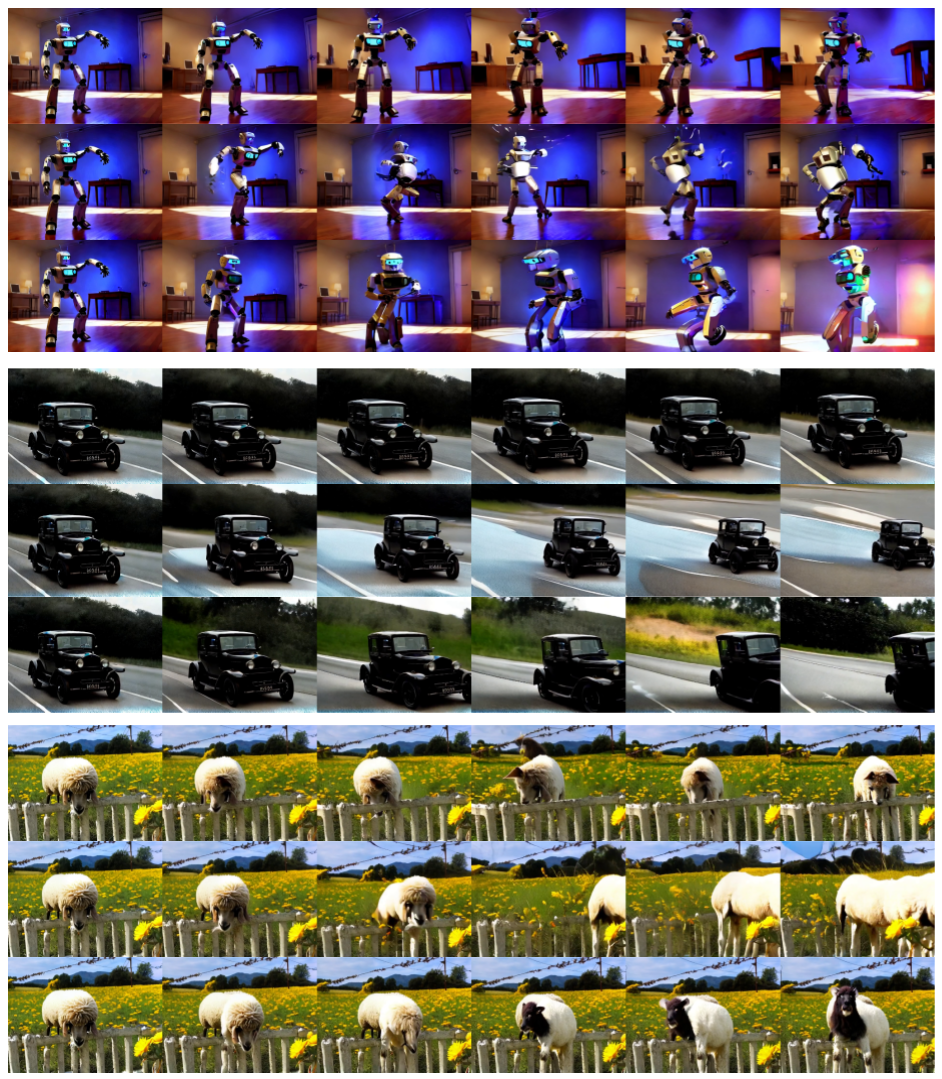}
    \caption{
        \small
        Examples of sampling results with the same conditioning input but different initial noises. Row 1-3 share same conditioning input, same for row 4-6, and row 7-9.
    }
    \label{fig:differet_init_noise}
\end{figure}

\subsection{Implementation Details}
We conduct extensive experiments on autoregressive chunk-by-chunk long video generation, with prompts from all categories in VBench~\cite{huang2024vbench}. We first use each prompt to generate a corresponding high quality guide image with StableDiffusion V1-5~\cite{rombach2022stablediffusion}, and use these images as initial guide images for performing autoregressive chunk-by-chunk video generation.
Additionally, for StableVideoDiffusion~\cite{blattmann2023stablevideodiffusion}, we use the EulerDiscreate scheduler~\cite{karras2022eulerdiscreate}. For ConsistI2V~\cite{ren2024consisti2v}, we use the DDIM scheduler~\cite{song2020ddim}. For OpenSoraPlan V1.3.0~\cite{yuan2024opensoraplan}, we use the EulerAncestralDiscrete scheduler~\cite{karras2022eulerdiscreate}. For CogVideoX~\cite{yang2024cogvideox}, we use the CogVideoXDPM scheduler~\cite{yang2024cogvideox, lu2022dpmsolver}. All these schedulers are default and recommended by their official implementations. Since these advanced schedulers introduce small amount of random noise during each step sampling, we set manual seed before every $k$-step and full-step sampling. Lastly, we empirically set the number of steps for full denoising to be 50, $k=8$, and the number of chunks to be 5 for all the main experiments unless otherwise stated. The size of each chunk follows the default chunk length used by each model -- StableVideoDiffusion: 25, OpenSoraPlan: 93, ConsistI2V: 16, CogVideoX: 49. 
We set video resolution (height by width) to $384\times512$ for StableVideoDiffusion, 
$480\times720$ for CogVideoX and OpenSoraPlanV1.3.0, $256\times256$ for ConsistI2V. The sampling speeds for each model are -- StableVideoDiffusion: 0.37s/step, OpenSoraPlan: 4s/step, ConsistI2V: 0.4s/step, CogVideoX: 3.8s/step. Experiments are conducted on a single Nvidia H100 GPU. Here, the sampling speeds highlight the importance of investigating applying chunk-by-chunk to the more inferior models -- by leveraging $k$-step to elevate their performance on long video generation, these models bring with them an advantage of much faster speed.

\begin{table*}
\small
\centering
\footnotesize

\scalebox{0.9}{

\begin{tabular}{lccccc}
\toprule
{\textbf{Methods/Metrics}} &  
\multicolumn{1}{c}{\textbf{Subject Consistency$\uparrow$}} &  \multicolumn{1}{c}{\textbf{Background Consistency$\uparrow$}}  &  \multicolumn{1}{c}{\textbf{Temporal Flickering$\uparrow$}} & \multicolumn{1}{c}{\textbf{Motion Smoothness$\uparrow$}} & \multicolumn{1}{c}{\textbf{Aesthetic Quality$\uparrow$}} \\


\cmidrule(lr){1-1} \cmidrule(lr){2-6} 

StableVideoDiffusion Baseline & 0.7707 & 0.8669 & 0.8560 & 0.8994 
        & 0.4370 \\
        StableVideoDiffusion $k$-step & 0.8209 & 0.8921 & 0.8719 & 0.9164 
        & 0.4702\\
    \cmidrule(lr){1-1} \cmidrule(lr){2-6} 
        ConsistI2V Baseline & 0.8559 & 0.9189 & 0.9611 & 0.9742 
        & 0.4789\\
        ConsistI2V $k$-step & 0.8956 & 0.9394 & 0.9646 & 0.9768 
        & 0.4872\\
    \cmidrule(lr){1-1} \cmidrule(lr){2-6} 
    OpenSoraPlanV1.3.0 Baseline & 0.8692 & 0.9318 & 0.9750 & 0.9872 
    & 0.5307 \\
    OpenSoraPlanV1.3.0 $k$-step & 0.8725 & 0.9329 & 0.9754 & 0.9874 
    & 0.5336\\
    \cmidrule(lr){1-1} \cmidrule(lr){2-6} 
    CogVideoX Baseline & 0.9703 & 0.9782 & 0.9816 & 0.9867 
    & 0.5355
    \\
    CogVideoX $k$-step & 0.9747 & 0.9798 & 0.9815 & 0.9873 
    & 0.5651
    \\
    \cmidrule(lr){1-1} \cmidrule(lr){2-6} 
    FIFO-Diffusion & 0.8436 & 0.9068 & 0.9215 & 0.9525 
    & 0.5473
    \\

\bottomrule

\end{tabular}

}

\caption{\footnotesize
    Metrics in VBench for Image-to-video task. StableVideoDiffusion and ConsistI2V are implemented with PseudoUNet3D, while OpenSoraPlanV1.3.0 and CogVideoX-5B-I2V are implemented with more complex 3D-DiT with much more parameters. We observed that the $k$-step search method have better improvement in UNet based small models than DiT based large models. Empirically, we used $k=8$ for all these experiments. 
    FIFO is not a chunk-by-chunk generation pipeline, and it is based on a T2V model, but we include it here due to its ability to also generate long videos.
}
\label{tab:vbench_metric}

\end{table*}

\begin{table*}
\small
\centering
\footnotesize

\scalebox{0.756}{

\begin{tabular}{lcccccccccccccccc}
\toprule
{\textbf{Methods/Metrics}} &  
\multicolumn{4}{c}{\textbf{Subject Consistency$\uparrow$}} &  \multicolumn{4}{c}{\textbf{Background Consistency$\uparrow$}}  &  \multicolumn{4}{c}{\textbf{Temporal Flickering$\uparrow$}} & \multicolumn{4}{c}{\textbf{Motion Smoothness$\uparrow$}} \\
\midrule
Number of Chunks   &     5 &  10 & 15 & 20 &   5 &  10 & 15 & 20  &   5 &  10 & 15 & 20 &   5 &  10 & 15 & 20    \\
\cmidrule(lr){1-1} \cmidrule(lr){2-5} \cmidrule(lr){6-9} \cmidrule(lr){10-13} \cmidrule(lr){14-17}
StableVideoDiffusion Baseline
& 0.7784 & 0.6613 & 0.6199 & 0.6003
& 0.8675 & 0.8115 & 0.7920 & 0.7830
& 0.8620 & 0.8011 & 0.7862 & 0.7782 
& 0.9042 & 0.8369 & 0.8180 & 0.8096 \\

StableVideoDiffusion $k$-step  
& 0.8280 & 0.7257 & 0.6783 & 0.6505 
& 0.8967 & 0.8374 & 0.8138 & 0.8009 
& 0.8798 & 0.8416 & 0.8189 & 0.8062 
& 0.9204 & 0.8814 & 0.8586 & 0.8459 \\

\cmidrule(lr){1-1} \cmidrule(lr){2-5} \cmidrule(lr){6-9} \cmidrule(lr){10-13} \cmidrule(lr){14-17}

ConsistI2V Baseline
& 0.9130 & 0.8708 & 0.8410 & 0.8069 
& 0.9475 & 0.9217 & 0.9014 & 0.8836 
& 0.9645 & 0.9652 & 0.9671 & 0.9693 
& 0.9752 & 0.9762 & 0.9778 & 0.9793\\

ConsistI2V $k$-step 
& 0.9318 & 0.8965 & 0.8543 & 0.8208 
& 0.9479 & 0.9219 & 0.9017 & 0.8864 
& 0.9645 & 0.9659 & 0.9683 & 0.9705 
& 0.9761 & 0.9767 & 0.9784 & 0.9800\\

\cmidrule(lr){1-1} \cmidrule(lr){2-5} \cmidrule(lr){6-9} \cmidrule(lr){10-13} \cmidrule(lr){14-17}

OpenSoraPlanV1.3.0 Baseline 
& 0.8692 & 0.7953 & 0.7582 & 0.7401 
& 0.9296 & 0.8933 & 0.8717 & 0.8589 
& 0.9721 & 0.9738 & 0.9762 & 0.9774 
& 0.9895 & 0.9895 & 0.9902 & 0.9894 \\

OpenSoraPlanV1.3.0 $k$-step  
& 0.8836 & 0.8248 & 0.7692 & 0.7397 
& 0.9277 & 0.8955 & 0.8648 & 0.8503 
& 0.9722 & 0.9729 & 0.9747 & 0.9752 
& 0.9897 & 0.9896 & 0.9901 & 0.9897\\

\cmidrule(lr){1-1} \cmidrule(lr){2-5} \cmidrule(lr){6-9} \cmidrule(lr){10-13} \cmidrule(lr){14-17}

CogVideoX Baseline
& 0.9548 & 0.9533 & 0.9329 & 0.9104 
& 0.9741 & 0.9629 & 0.9489 & 0.9350 
& 0.9876 & 0.9879 & 0.9880 & 0.9877 
& 0.9935 & 0.9935 & 0.9934 & 0.9933 \\

CogVideoX $k$-step  
& 0.9677 & 0.9535 & 0.9373 & 0.9198 
& 0.9754 & 0.9698 & 0.9616 & 0.9505 
& 0.9899 & 0.9914 & 0.9910 & 0.9905 
& 0.9941 & 0.9947 & 0.9945 & 0.9942 \\

\bottomrule

\end{tabular}

}

\caption{\footnotesize
Results from longer video generation with each method. Smaller models such as StableVideoDiffusion and ConsistI2V benefit more from $k$-step search than larger models like OpenSoraPlanV1.3.0 and CogVideoX do.
}
\label{tab:ablation_chunk}
\end{table*}

\subsection{Impact of Initial Noise on Output Quality}
Gaussian initial noise introduces diversity and stochasticity to diffusion models. For an I2V model, even when a strong guide image is provided to constrain the overall semantic content of the output, the initial noise still has a significant impact on the final output. For example, as shown in Figure~\ref{fig:differet_init_noise}, there are noticeable differences in video qualities depending on the initial noises.
We also present quantitative results to highlight the influence of the initial noises on each model. In our experiments, we randomly sampled 10 different initial noises for each conditioning input, allowing models to denoise from these varied starting points. We then evaluated the output videos using VBench metrics. The results show that even with the same conditioning input, different initial noises can lead to substantial variations in video quality (See Table~\ref{tab:noise_analysis}). 

\subsection{Evaluation Metrics}

To evaluate the quality of the generated videos, we selected five dimensions from VBench's~\cite{huang2024vbench} Video Quality category including Subject Consistency, Background Consistency, Temporal Flickering, Motion Smoothness, and Aesthetic Quality. Specifically, \textbf{Subject Consistency} is defined as the average cosine similarity between the DINO~\cite{caron2021dino} features of every pair of consecutive frames; \textbf{Background Consistency} is similar to Subject Consistency except that the DINO features are replaced with CLIP~\cite{radford2021clip} features; \textbf{Temporal Flickering} is defined as the mean absolute difference across frames in pixel space. As for \textbf{Motion Smoothness}, a pretrained video frame interpolation model AMT~\cite{li2023amt} is first utilized to reconstruct all the odd-number-frames from the even-number-frames, after which a Mean Absolute Error is calculated between the ground-truth and reconstructed odd-number-frames.
\textbf{Aesthetic Quality}, reflecting frame-level aesthetic aspects, is calculated with the LAION aesthetic predictor~\cite{LAIONaes} across all frames. We observed significant improvement for StableVideoDiffusion and ConsistI2V (See Table~\ref{tab:vbench_metric}). For larger models such as OpenSoraPlanV1.3.0, and CogVideoX, the improvement brought by $k$-step search is not very significant because the base models are already very robust to initial noises. In these cases, naive chunk-by-chunk would likely suffice.

\subsection{Ablation Studies}

\subsubsection{Number of Chunks}
We further conducted experiments comparing the effect of varying the number of chunks across the different models in Table~\ref{tab:ablation_chunk} based on the VBench metrics. StableVideoDiffusion continues to benefit greatly from $k$-step selection even up to as long as 20 chunks. OpenSoraPlan shows strong performance itself, but can really benefit from $k$-step for chunks of 5, 10 and 15. ConsistI2V also benefitted greatly from $k$-step across the board. CogVideoX is the strongest by itself, with $k$-step bringing just marginal benefits.

\begin{figure}
    \centering
    \includegraphics[width=0.5\textwidth]{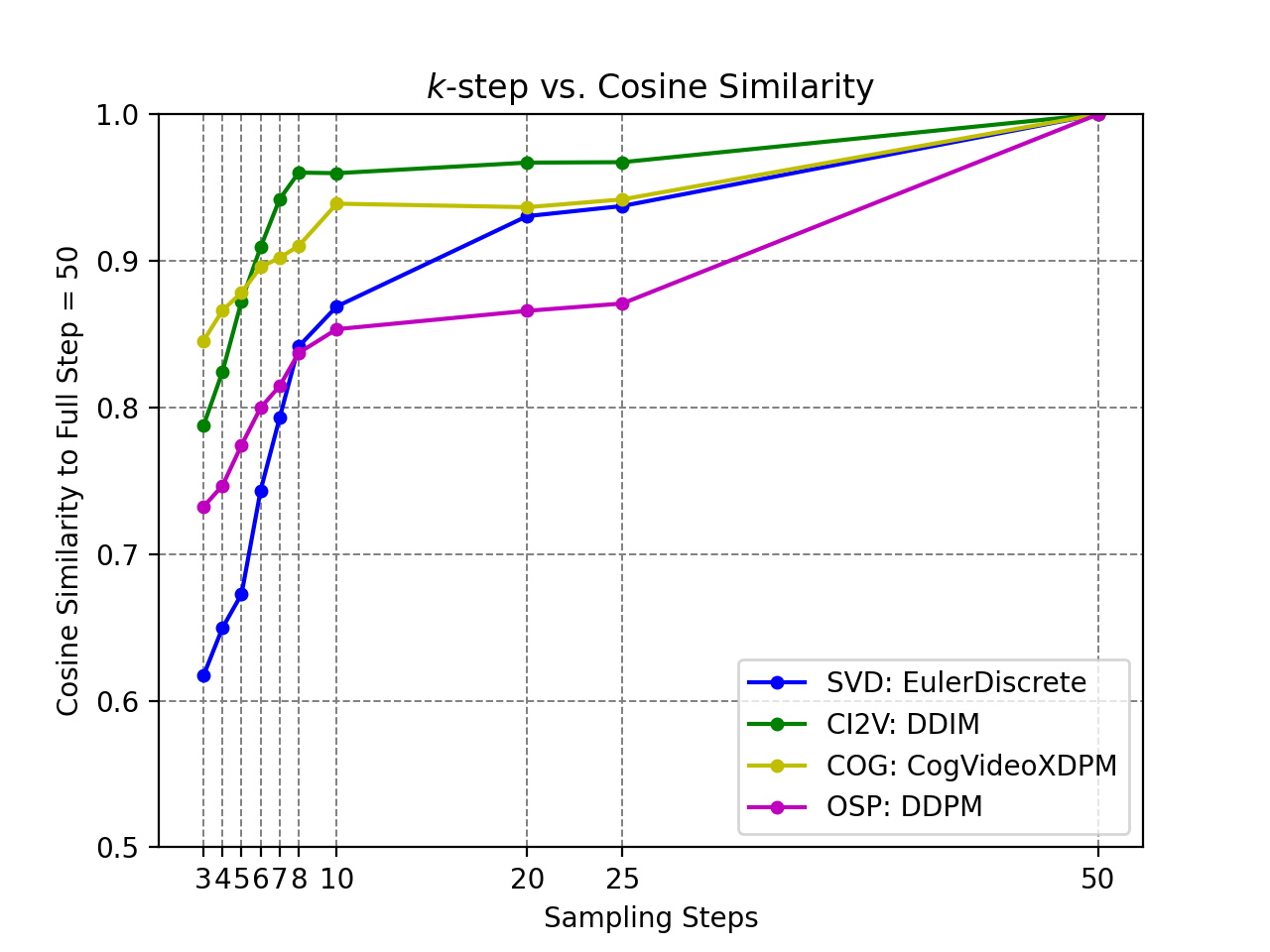}
    \caption{\footnotesize
    For each model, we apply its recommended scheduler and calculate the cosine similarity between a video generated after $k$ steps and a video generated with 50 steps with the same noise and conditioning inputs.
    }
    \label{graph:kstep_ablation}
\end{figure}

\subsubsection{Choice of $k$-step}
Most diffusion models adapt the predefined beta schedule introduced in the DDPM~\cite{ho2020denoising} paper, with a 1000 sampling steps. During inference, the number of sampling steps is
typically reduced to between 25 to 100, due to the ability of advanced sampling scheduler such as PNDM~\cite{liu2022pndm}, DPMSolver~\cite{lu2022dpmsolver}, and EulerAncestralDiscrete~\cite{karras2022eulerdiscreate}. Furthermore, if we only need to make a quick judgment whether an initial noise will lead to a satisfactory output, we only need to do inference from that noise for a rather small $k$ steps. 



We conducted experiments on the relation between video quality and $k$ (see Figure~\ref{graph:kstep_ablation}) and found that when $k$ is small, by increasing $k$, the video quality increases quickly. However, after reaching a threshold, adding more sampling steps becomes less cost efficient. Empirically, we selected $k=8$ as a good balance between an acceptable video quality and a reasonable cost.

\section{Conclusion} 

This work demonstrates the feasibility of long video generation in a chunk-by-chunk manner based on pretrained I2V models. We show that the initial noise plays an important role in generating a video, subject to the model and condition. We also provide a quick $k$-step search pipeline to choose a good noise from multiple noise candidates, mitigating the degradation effect of naive chunk-by-chunk video generation by smaller I2V models. We also found that larger I2V models surprisingly demonstrated strong robustness against error accumulation, and naive chunk-by-chunk long video generation may suffice here. 

\subsection*{Limitations and Future Works}
Although our method is able to quickly choose a good noise from multiple candidates, it still relies on random sampling instead of explicit refining. A more desirable approach would be to refine an arbitrary initial noise, which we will delegate to future work. Another possible future work is to introduce degradation such as blurring and noising (in pixel space) to conditioning inputs during training, in order to strengthen a base I2V model against common degradations during inference. Most importantly, while our work has allowed us to create videos up to 20 chunks or more at a high level of quality, we would not be able to confidently claim that we can generate videos of infinite length. This is because even with $k$-step, small amount of errors will still be introduced, which will eventually result in unrecoverable amount of error after many chunks. A potential foolproof approach would thus be to develop an effective feedback loop, either automatically or with human in the loop.


{
    \small
    \bibliographystyle{ieeenat_fullname}
    \bibliography{main}
}


\end{document}


\title{Supplementary Materials For ``Towards Generating Long Videos Chunk by Chunk''}  

\twocolumn[{%
\renewcommand\twocolumn[1][]{#1}%
\maketitle
\begin{center}
    \vspace{-10pt}
    \includegraphics[
    width=1.0\textwidth,
    height=0.6\textwidth
    ]{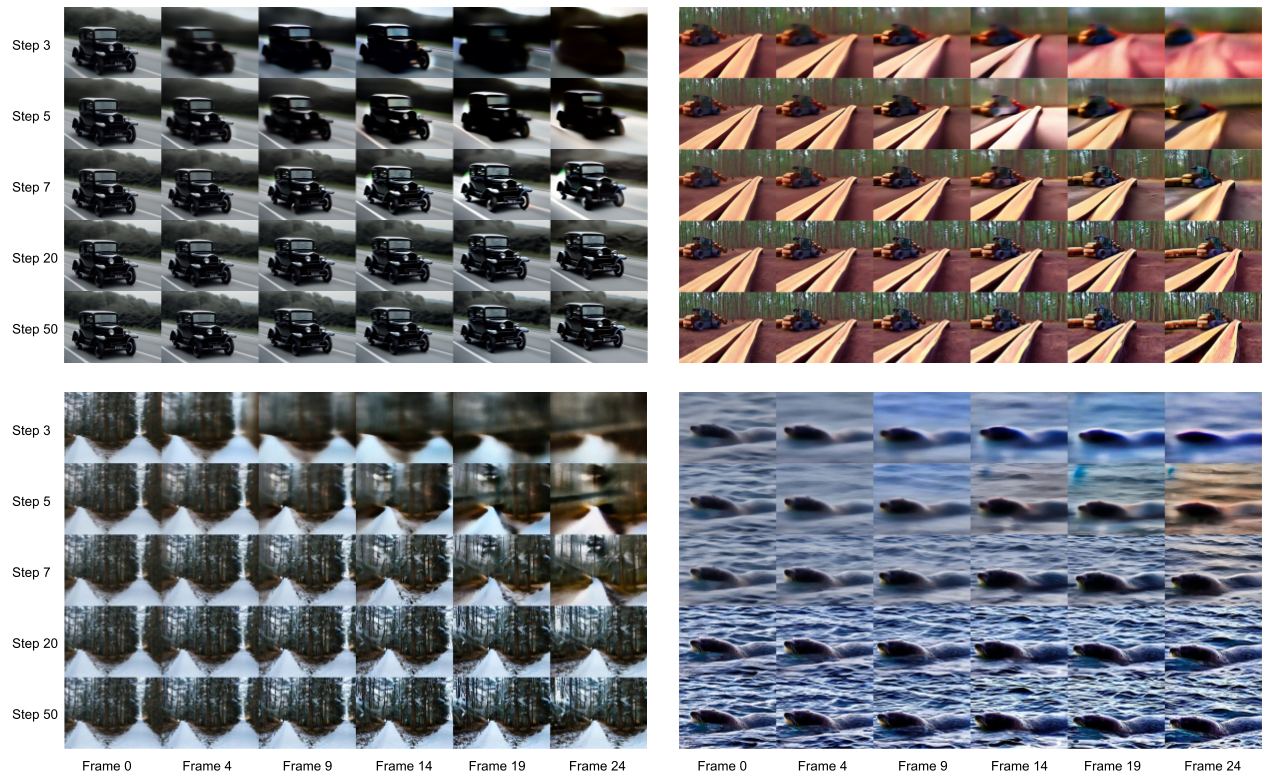}
\vspace{-0.5cm}
\captionof{figure}{
Qualitative results for same initial noise and conditioning input but different $k$-steps. Although fewer $k$-steps typically result in blurrier, lower-quality videos and more $k$-steps produce clearer, higher-quality outputs, they both maintain the same spatio-temporal semantics. We notice that the differences between step 20 and step 50 are barely noticeable, and step 7 is sufficient to reflect the quality of step 50.
}
    \vspace{3pt}
    \label{supfig:kstep}
    \end{center}%
}]

\begin{table*}
\small
\centering
\footnotesize

\scalebox{1.25}{

\begin{tabular}{lcccc}
\toprule
{\textbf{Configurations/Methods}} &  
\multicolumn{1}{c}{\textbf{StableVideoDiffusion}} &  \multicolumn{1}{c}{\textbf{ConsistI2V}}  &  \multicolumn{1}{c}{\textbf{OpenSoraPlanV1.3.0}} & \multicolumn{1}{c}{\textbf{CogVideoX}} \\

\cmidrule(lr){1-1} \cmidrule(lr){2-5} 

Denoising Model Backbone & UNet-3D & UNet-3D & DiT & DiT \\
VAE & VAE-2D & VAE-2D & VFVAE & Causal VAE-3D \\
Text Encoder & - & CLIP & mT5 & T5 \\
Number of Parameters & 1.5 B & 1.2 B & 2.7 B & 5.5 B \\
Resolution & $384\times 512$ & $256 \times 256$ & $480 \times 720$ & $480 \times 720$ \\
Chunk Length & 25 & 16 & 93 & 49 \\
Context-free Guidance & 3.0 & 7.5 & 6.0 & 7.5\\
Sampling Step & 50 & 50 & 50 & 50 \\
$k$-step & 8 & 8 & 8 & 8 \\
Number of Candidates & 5 & 5 & 5 & 5 \\
Sampling Speed(s/step) & 0.37 & 0.4 & 4.0 &
3.7 \\
Sampling Speed(s/frame) & 1.33 & 2.25 & 3.87 & 6.79\\
Scheduler & EulerDiscrete & DDIM & EulerAncestralDiscrete & CogVideoXDPM \\
Data Type & float16 & float16 & float16 & float16 \\
Inference Memory Usage & 9 GB & 10 GB & 31 GB & 27 GB \\

\bottomrule

\end{tabular}

}

\caption{
    Configuration details for models we used for experiments.
}
\label{tab:implementation_detail}

\end{table*}

\thispagestyle{empty}
\appendix


\section{Experiment Details}

In Table~\ref{tab:implementation_detail}, we list configuration details for our experiments in long video generation with $k$-step search, on StableVideoDiffusion~\cite{blattmann2023stablevideodiffusion}, ConsistI2V~\cite{ren2024consisti2v}, OpenSoraPlanV1.3.0~\cite{yuan2024opensoraplan}, CogVideoX~\cite{yang2024cogvideox}. All our experiments are conducted on an Nvidia H100 GPU with 80 GB memory.

\textbf{StableVideoDiffusion} leverages a pretrained 2D UNet~\cite{rombach2022stablediffusion, ronneberger2015unet} and incorporates additional modules specifically designed to handle motion semantics. The model processes input by concatenating the noisy latent and the guide image latent along the channel dimension, introducing extra input channels to the first convolutional layer to accommodate this setup. Furthermore, StableVideoDiffusion employs a hybrid conditioning mechanism that combines concatenation and cross-attention techniques. In the original UNet designed for text-to-image (T2I) tasks, text embeddings serve as conditioning inputs. However, StableVideoDiffusion adapts these modules to utilize CLIP image embeddings~\cite{radford2021clip} instead. Since CLIP operates within a shared latent space for both image and text modalities, the pretrained cross-attention weights in the UNet remain effective, facilitating adaptation to the video synthesis task while maintaining compatibility with motion and image semantics. \textbf{ConsistI2V} employs FrameInit~\cite{ren2024consisti2v} for noise initialization, leveraging the low-frequency components of the guide frame to enhance the stability of video generation. Additionally, ConsistI2V utilizes concatenation as its conditioning mechanism by introducing first-frame condition injection into the temporal dimension by replacing the noise of the first frame with the guide image latent, ensuring a stable and consistent condition for video synthesis. These two model are rather small and fast, but likely to generate some artifacts, which will be magnified over chunks. 

\textbf{OpenSoraPlanV1.3.0} and \textbf{CogVideoX} both apply Diffusion Transformer (DiT)~\cite{peebles2023dit} as their denoising backbone, and employ masked conditioning by concatenating the noisy video latent, the mask and the masked video all along the channel dimension, providing structured guidance to the diffusion transformer for improved generation quality and temporal coherence. Also, they utilize larger text encoders T5~\cite{raffel2020t5} and mT5~\cite{xue2020mt5} to accommodate longer and more detailed text descriptions as extra condition for Image-to-Video (I2V) generation. These model are also trained in delicately curated datasets in a large scale, and thus shows more robustness against initial noise and artifacts.

\section{$k$-step Qualitative Results}
An essential assumption underlying our $k$-step search methodology is that, with fixed initial noise and conditional inputs, the model is expected to generate videos that preserve consistent spatio-temporal semantic content across different sampling steps, despite of variations in the quality in terms of local details.
Although the experimental section provides similarity curves illustrating the relationship between videos generated with different $k$-step and corresponding 50-step fully denoised videos, this section seeks to present a more qualitative demonstration (See Figure~\ref{supfig:kstep}).

\section{More Qualitative Results in Long Video}
With our $k$-step search method, we successfully extend the capability of base Image-to-Video (I2V) models to generate long videos. For a detailed visualization, please refer to the local \textit{\href{./gallery.html}{gallery.html}} file included in the supplementary folder, where we provide examples of long videos generated using the $k$-step search approach. Note that, for webpage presentation purposes, all videos have been resized to a resolution of $480 \times 640$.

{
    \small
    \bibliographystyle{ieeenat_fullname}
    \bibliography{main}
}